\documentclass[conference]{IEEEtran}
\IEEEoverridecommandlockouts

\usepackage{placeins}
\usepackage{array} 
\usepackage{booktabs} 
\usepackage{ragged2e} 

\usepackage{float}
\usepackage{multirow} 
\usepackage{booktabs} 
\usepackage{amsmath} 
\usepackage{cite}
\usepackage{amsmath,amssymb,amsfonts}
\usepackage{algorithmic}
\usepackage{graphicx}
\usepackage{textcomp}
\usepackage{xcolor}
\usepackage{subcaption}
\usepackage{cuted}     
\usepackage{caption}   
\def\BibTeX{{\rm B\kern-.05em{\sc i\kern-.025em b}\kern-.08em
    T\kern-.1667em\lower.7ex\hbox{E}\kern-.125emX}}

\begin{document}
\nocite{b2,b3,b7,b9,b10,b14,b16,b18,b19,b22,b23,b24,pnas,mypaper2,mypaper3}
\title{DWI Connectome-based Sex Classification in Young using Supervised Variational Autoencoder\\

}

\author{
\IEEEauthorblockN{
\begin{minipage}[t]{0.33\linewidth} 
\centering
Aditya Kothapalli\\
\textit{Dept. of Computer Science and Engineering}\\
\textit{Amrita School of Computing}\\
\textit{Amrita Vishwa Vidyapeetham}\\
Bengaluru, India\\
bl.en.u4aie23161@bl.students.\\amrita.edu
\end{minipage}
\hfill 
\begin{minipage}[t]{0.33\linewidth} 
\centering
Aaryan Gera\\
\textit{Dept. of Computer Science and Engineering}\\
\textit{Amrita School of Computing}\\
\textit{Amrita Vishwa Vidyapeetham}\\
Bengaluru, India\\
bl.en.u4aie23102@bl.students.\\amrita.edu
\end{minipage}
\hfill 
\begin{minipage}[t]{0.33\linewidth} 
\centering
Sathvik Reddy V\\
\textit{Dept. of Computer Science and Engineering}\\
\textit{Amrita School of Computing}\\
\textit{Amrita Vishwa Vidyapeetham}\\
Bengaluru, India\\
bl.en.u4aie23127@bl.students.\\amrita.edu
\end{minipage}
}\\ 
\IEEEauthorblockN{
\begin{minipage}[t]{0.33\linewidth} 
\centering
Ninad Aithal\\
\textit{Vision and Artificial Intelligence Lab}\\
\textit{Indian Institute of Science (IISc)}\\
Bengaluru, India \\
ninadaithal@iisc.ac.in
\end{minipage}
\hfill 
\begin{minipage}[t]{0.33\linewidth} 
\centering
Debanjali Bhattacharya*\\
\textit{Dept. of Artificial Intelligence}\\
\textit{Amrita School of Artificial Intelligence}\\
\textit{Amrita Vishwa Vidyapeetham}\\
Bengaluru, India\\
b\_debanjali@blr.amrita.edu
\end{minipage}
\hfill 
\begin{minipage}[t]{0.33\linewidth} 
\centering
Neelam Sinha\\
\textit{Centre for Brain Research}\\
\textit{Indian Institute of Science (IISc)}\\
Bengaluru, India \\
neelam@cbr-iisc.ac.in
\end{minipage}
}
}

\maketitle

\begin{abstract}

During adolescence, significant changes in brain development are influenced by hormonal changes that impact both the body and the brain. This study presents sex classification in 787 young adults having age range of 15-25 years using Diffusion-weighted (DWI) structural connectome, focusing on 82 Desikan-Killiany atlas ROIs. A supervised variational autoencoder (SVAE) is used to extract latent space features from DWI connectome data. Although several studies explored differences in brain structure and function between sexes, the main contribution of this study lies in incorporating SVAE to learn a compact, regularized latent representation for the task of sex classification. Initial exploratory data analysis shows that these features can capture the differences in sex-specific latent space patterns. Subsequently, the learned latent space SVAE features are fed to different ML models for sex classification. Among the considered ML classifiers, the multi-layer perceptron yields the highest accuracy of 80\% in classifying sex. This demonstrates the effectiveness of SVAE in learning compact, informative representations from DWI connectome for sex classification. The present work highlights the potential of using DWI brain connectomics and deep learning to differentiate sex, paving the way for future research on diverse populations that may include transgender individuals, to better understand sex-related variations in brain structure.

\end{abstract}
\medskip

\section{Introduction}

Neuroscience research and clinical studies require fundamental knowledge about how sex affects brain structure and operation in the human body.  
In past decades, several studies have been performed for sex classification utilizing multimodal medical images \cite{b2,b9,b14,b16,b22,b23,b24,pnas}. There is a growing interest in understanding how male and female brains differ structurally and functionally, as the literature highlights adaptive complementarity between males and females across various cognitive, behavioral, and neurological domains \cite{pnas,b22,b23,b24}. While MRI-based classification is well-studied in different medical image classification scenario \cite{b3,b7,b10,mypaper2,mypaper3,b18,b19}, few studies have explored DWI structural connectome for the task of sex classification. Understanding and classifying biological sex using DWI connectome can provide valuable insights into how male and female brains' structural connectivity is organized differently. 
The proposed study develops a deep learning framework for sex classification using brain connectomes obtained from DWI structural connectivity. The research investigates the applicability of supervised variational autoencoder (SVAE) to learn a discriminative, low-dimensional representation of the connectome data. The goal of SVAE is to learn a generative model while incorporating label information to guide the learning process, specifically for classification tasks. By this method, efficient classification can be done once a compact latent space is built from a high-dimensional brain connectome matrix. Clinically, this approach may facilitate the study of neurodevelopmental and neuropsychiatric disorders that show sex-based differences, thereby affecting their emergence, development, and or therapeutic measures.

\begin{figure*}[btp]
    \centering
    \includegraphics[width=1\linewidth]{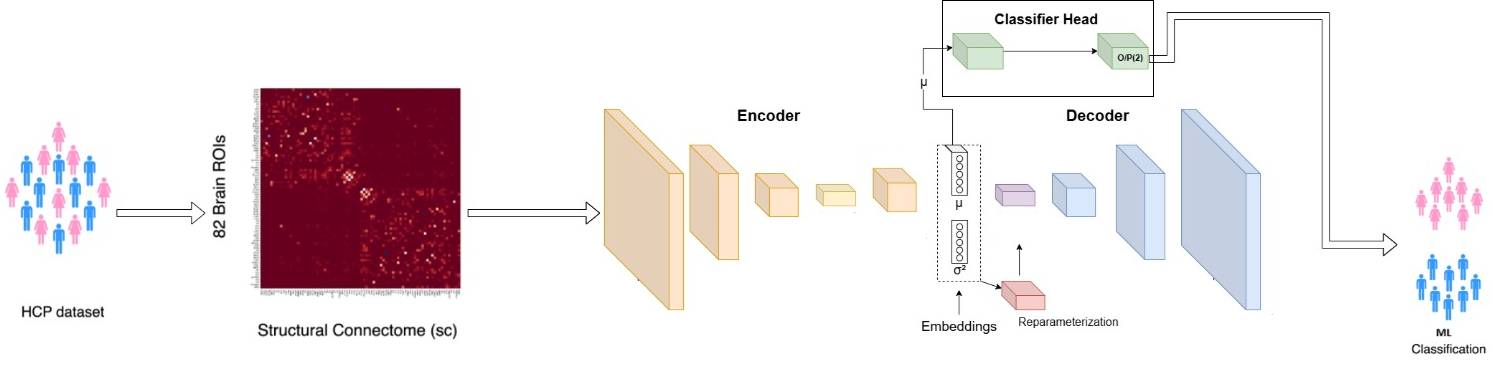}
    \caption{Schematic overview of the proposed approach for sex classification}
    \label{fig:bdVAE}
\end{figure*}

The main contribution of this study is to analyze the characteristics of the learned SVAE latent space features to visually and quantitatively assess their ability to distinguish between male and female subjects. To the best knowledge of authors, there is no other study that investigate the potential of DWI-derived structural connectome using SVAE latent space embeddings for classifying sex of the individual.

\section{Related Works}

In last few decades several studies have performed sex classification using multi-modal medical images.
S. L. Mendes et al.\cite{b2} presents an integrative study that employs a 3D-CNN framework to estimate sex, age and psychiatric  status from conventional MRI of children as well as young, addressing critical need for robust biological biomarkers in neuropsychiatric research.
P. Baldinger‐Melich et al.\cite{b3} presents an original investigation into the  neuroanatomical underpinnings of gender and gender differences using a multivariate pattern analysis approach on structural MRI.
Weis et al.\cite{b9} explores the use of resting-state fMRI connectivity for distinguishing between male and female brains. The study also found that gender-related differences in connectivity were particularly pronounced in regions like DMN network, limbic system, and visual areas. These findings suggest the importance of intrinsic brain networks in encoding gender-specific traits, even in the absence of explicit tasks or external stimuli. The paper supports the notion that machine learning methods applied to rs-fMRI connectivity data can serve as effective tools for gender classification, revealing consistent patterns that generalize well across individuals.
Madhura I. et.al in paper\cite{pnas} examined sex differences using structural connectome where they have found that male and female brains are optimized for intra-hemispheric and inter-hemispheric communications, respectively.
Antonia N. Kaczkurkin et.al \cite{b23} reviewed various studies on sex differences in brain structure specially that associated with different psychopathology like anxiety, psychosis, depression, and attention-deficit or hyperactivity symptoms. 
Matthis Ebel et.al.\cite{b22} reported how precisely sex is differentiated from brain GM using ML classifiers. 

Contrary to this literature, in this paper we proposed a different approach that studied SVAE latent space embeddings of DWI structural connectome for sex classification. More specifically, we have addressed the research question, \textit{how effective are SVAE latent space features in capturing sex-specific differences using DWI connectome?} 

The remaining sections of this paper are structured as follows: Section~\ref{dataset} provides the data description, followed by the details of proposed SVAE methodology which is described in Section~\ref{AA}. The result and analysis of the study is demonstrated in Section~\ref{results}, followed by conclusion in Section~\ref{conclusion}. 

\section{Dataset Description}
\label{dataset}

We have utilized DWI MRI images from the Human Connectome Project HCP\footnote{http://www.humanconnectomeproject.org/}.  A total of $787$ healthy subjects (age range of $15-25$ years) with $351$ male and $436$ females are considered in this study. For each individual subject, diffusion data includes diffusion weighting, direction (AP), time series, brain mask, and gradient nonlinearity data.

\section{Methodology}
\label{AA}

The present study utilizes SVAE to capture the structural connectivity patterns that distinguish male and female brain. Performance of SVAE is evaluated with different ML classifiers on latent space embedding features. The overview of present study is illustrated in Figure~\ref{fig:bdVAE}.

\subsection{DWI Whole-Brain Probabilistic Tractography}

Structural MRI preprocessing is performed using the Freesurfer (version 7.1.0), which parcellates the cerebral cortex into gyri and sulci and offers 82 brain regions (ROIs) based on Desikan-Killian Atlas or DK Atlas (refer \cite{mypaper2}). DWI data underwent initial preprocessing using the MRtrix toolbox, which included denoising and removal of Gibbs ringing artifacts. Further preprocessing stages are performed with FSL (version 6.0.7.7), including brain extraction using brain extraction tools. Motion correction, distortion correction due to induced susceptibility using topup, and eddy-current are also executed.

MRtrix3 is utilized further to perform anatomically-constrained probabilistic tractography. This process began with five-tissue-type segmentation, which differentiates cortical and subcortical GM, WM, CSF, and other potential pathological tissues (if any). Next, voxel-wise fibre orientation densities were estimated using the constrained spherical deconvolution algorithm. This allows the modeling of complex fibre crossings within each voxel. MRtrix’s ability to resolve multiple crossing fibres within a single voxel enabled a comprehensive decomposition of the diffusion signal across various directions.
Tractography analysis for whole-brain is subsequently performed by seeding streamlines at the gray matter–white matter interface, initiating the propagation of streamlines that represent WM pathways throughout the brain. For each participant, 10 million streamlines were generated. The default parameters included a maximum streamline length of 250 mm and an angular threshold of 0.06 radians, defining the streamline termination criteria and influencing seeding, propagation, and overall tractography.
Following tractography, connectome matrices are constructed to represent the number of streamlines connecting each pair of DK atlas ROIs. The tractography pipeline is executed using HPC; on a dual-CPU Intel® Xeon® Gold 6240 system at 2.60 GHz, equipped with 192 GB of RAM.

\begin{figure*}[btp]
    \centering
    \includegraphics[width=0.8\linewidth]{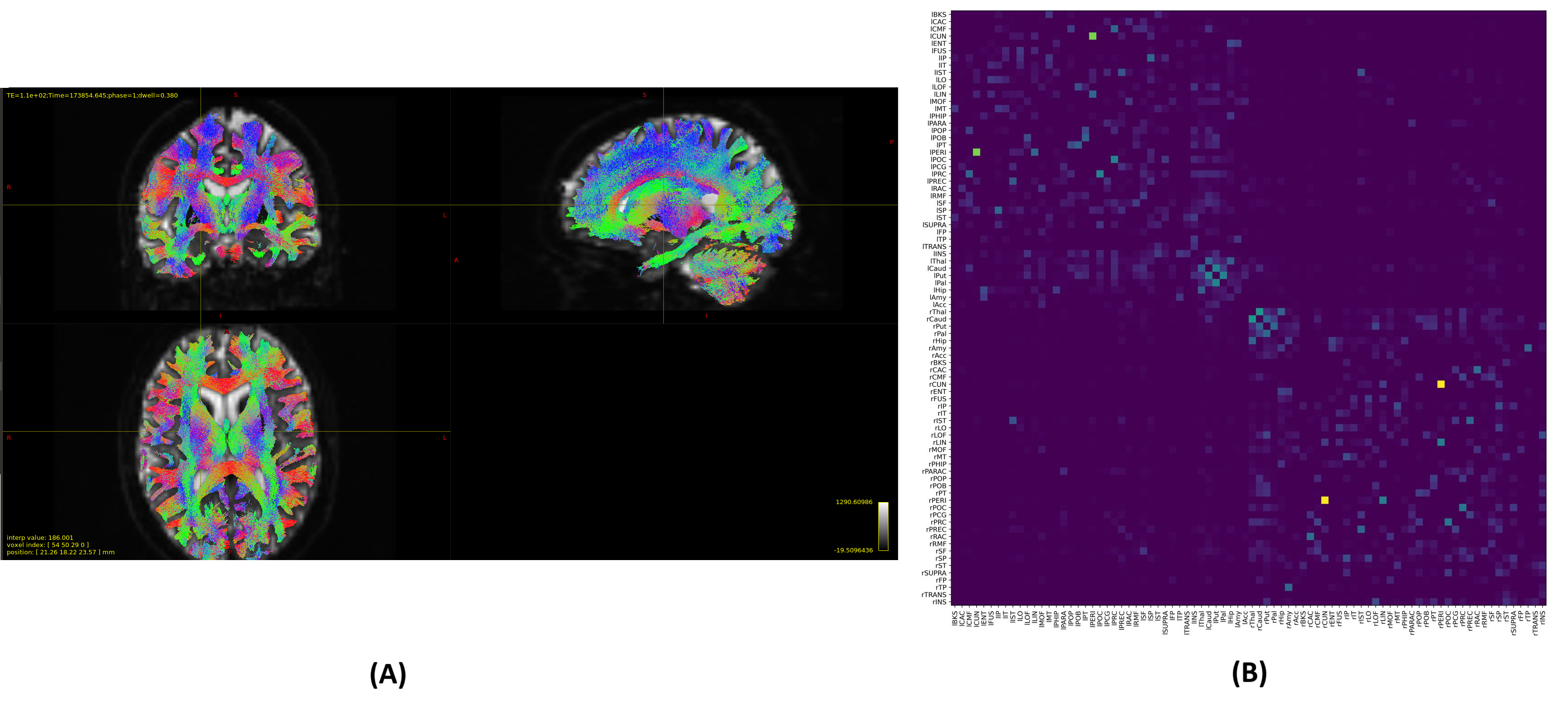}
    \caption{DWI Connectome for one representative subject. (A) The generated streamlines across different brain regions according to DK Atlas using MRTrix. (B) DWI connectome: The values in the DWI connectivity matrix represents the strengths of fibre connectivity across different brain regions. The diagonal and off diagonal values show intra-hemispheric and inter-hemispheric connectivity, respectively.}
    \label{fig:dwiconnectome}
\end{figure*}

\subsection{SVAE Architecture}

\begin{figure}[btp]
    \centering
    \includegraphics[width=1\linewidth]{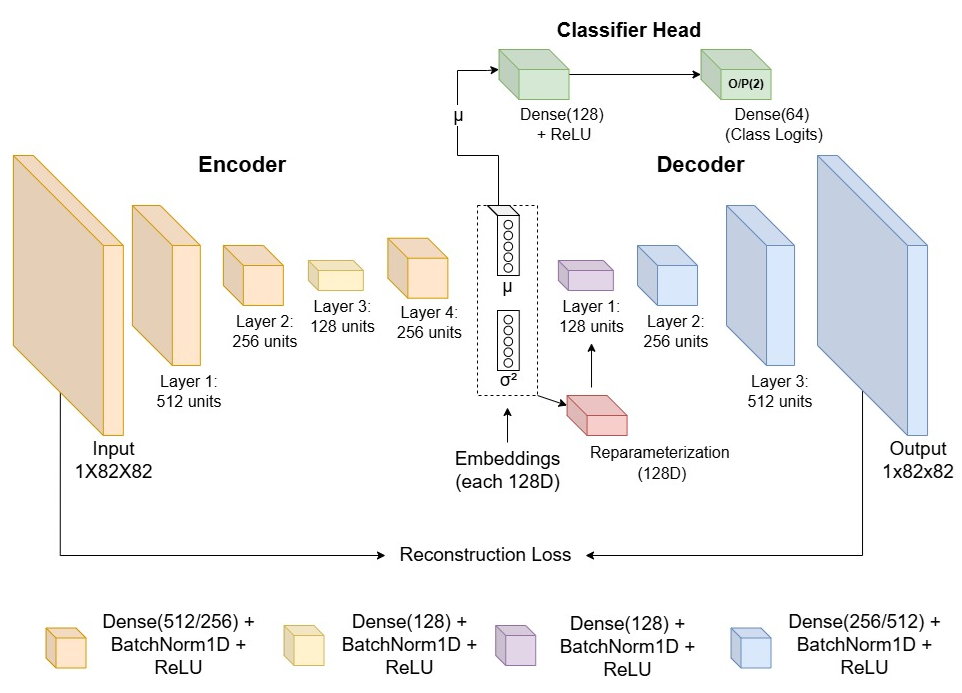}
    \caption{The detailed SVAE architecture used in this study.}
    \label{fig:SVAEarch}
\end{figure}

A SVAE architecture is developed to derive discriminative low-dimensional representations from the DWI connectome data to perform sex classification. The SVAE is implemented as a \texttt{PyTorch nn.Module}, integrated an encoder, decoder, and classifier head using fully connected layers. The proposed SVAE sex classification model is detailed in Figure~\ref{fig:SVAEarch}.
The encoder mapped DWI connectome features to a 128-dimensional latent space feature. It comprises three linear layers (512, 256, 128 dimensions), with each layer including by batch normalization and a ReLU activation function. 
The final encoder output, a 256-dimensional vector, is split into 128-dimensional $\mu$ and 128-dimensional $log\_var$ (log variance), representing mean and log-variance of Gaussian distribution in the latent space, respectively. The log-variance controls the spread or uncertainty around $\mu$. A higher variance indicates the SVAE is more uncertain about where to place this input in the latent space. This design reflects that SVAE learns to model the input as a probability distribution in latent space, rather than a single point, unlike traditional autoencoder model. This allows the SVAE to model variability in the data and generate smooth interpolations between points in the latent space.

The reparameterization trick in SVAE enables sampling from the latent space  \( z \) by transforming the mean and log-variance of the distribution from 
\( \mathcal{N}(\mu, \exp(\log\_var)) \) into a differentiable form. Typically, it is computed as:
\[
z = \mu + \epsilon \cdot \exp\left(0.5 \cdot \log\_var\right)
\]
where \( \epsilon \sim \mathcal{N}(0, I) \).

For numerical stability, log\_var is clamped between -10 and 10, enabling efficient backpropagation through the stochastic sampling step.
A classifier head is introduced in the VAE architecture in order regularize the SVAE latent space so that autoencoder model is not only useful for reconstruction but also makes the latent space more discriminative for classification. The classifier head branched from the $\mu$ vector, which is the location in the latent space that best represents the given input; mapping the 128-dimensional latent representation to a 64-dimensional hidden layer with ReLU activation, and gives as input to 2-dimensional logits for binary sex classification. The $\mu$ vector which represents the central tendency of the learned latent distribution of SVAE, served as the features for subsequent classification using ML models. 

To establish a baseline, the performance of a standard Convolutional Neural Network was also assessed. This baseline model achieved a classification accuracy approximately 6\% below that of the SVAE-derived features, highlighting the enhanced discriminative power of the latent representations learned by the SVAE and justifying its selection as the primary feature extraction method for this work.


The SVAE is trained end-to-end using a composite loss function:
\[
L_{\text{total}} = L_{\text{reconstruction}} + \beta * L_{\text{KL}} + \alpha * L_{\text{classification}}
\]
The reconstruction loss ($L_{\text{reconstruction}}$), calculated using the mean squared error (MSE), measured the difference between the original and reconstructed DWI features. The KL Divergence loss ($L_{\text{KL}}$), calculated as,
\[ L_{\text{KL}} = -0.5 * \sum_{i=1}^{d} (1 + \log\_var - \mu^2 - \exp(\log\_var))
\]
The KL Divergence loss regularized the latent space of dimension `d' to a standard normal distribution and is weighted by $\beta$=1.0. The classification uses cross-entropy loss function  ($L_{\text{classification}}$), encourages the model to learn discriminative features in the latent space which are useful for classification.
The model training is performed over 50 epochs using batch size of 32. The Adam optimizer is set at learning rate of $10^{-4}$. To ensure stability, gradient norm clipping with a maximum norm of 1.0 is applied. Training occurred on a CUDA-enabled GPU.

\subsection{ML Classification}

The 128-dimensional learned latent space features of SVAE is fed to different ML models for sex classification. A broad set of standard ML classifiers is used to thoroughly evaluate how the SVAE latent features perform for sex classification. These include SVM with RBF kernel, Logistic Regression, Random forest, Gradient-boost, XGBoost, LightGBM, Multi-layer Perceptron (MLP), K-nearest neighbors ($k=7$), Gaussian Naive Bayes, Linear Discriminant Analysis (LDA) and Ensemble methods like Voting Classifier (SVM, XGBoost, MLP with soft voting), Stacking Classifier (SVM, XGBoost as base, Logistic Regression as meta-estimator). For models sensitive to class imbalance (XGBoost, LightGBM, Logistic Regression, SVM, and ensembles containing them), \texttt{scale\_pos\_weight=1.24} and \texttt{class\_weight='balanced'} is used. A 10-fold stratified cross-validation strategy is employed. Within each fold, SVAE latent features are standardized using \texttt{StandardScaler} via a Scikit-learn Pipeline. The mean of different performance measures along with standard deviation across 10 folds are computed from the obtained confusion matrix in classifying sex.

\section{Results and Discussion}
\label{results}

\begin{figure}[btp]
    \centering
    \includegraphics[width=1\linewidth]{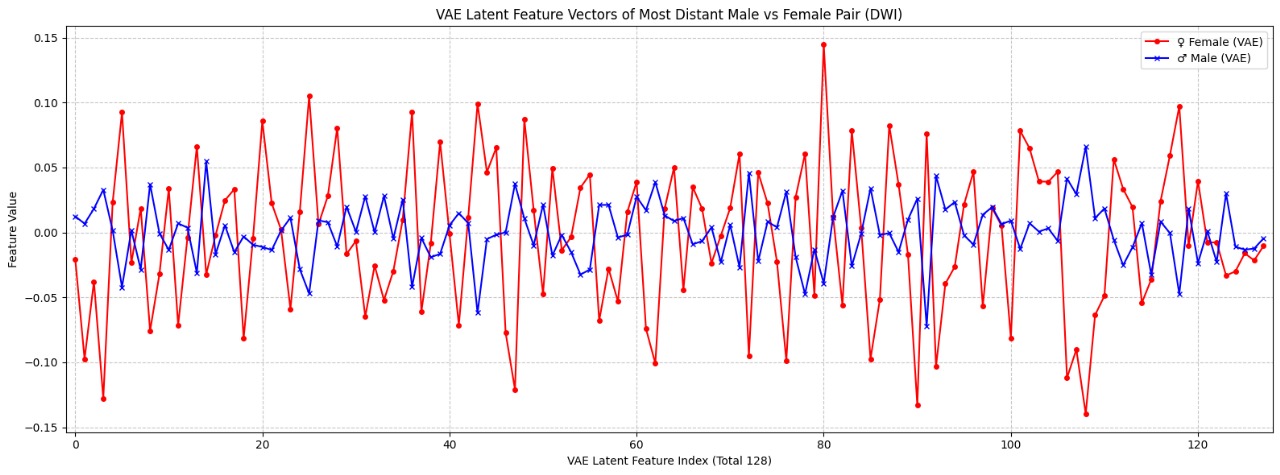}
    \caption{Illustration of SVAE latent space features for one representative male \textit{(blue)} and female \textit{(red)} subject. The plot clearly shows the differences in SVAE latent space pattern between male and female.}
    \label{fig:graph}
\end{figure}

The generated DWI brain connectome with 82 DK Atlas ROIs for one subject is shown in Figure~\ref{fig:dwiconnectome}. A preliminary visual analysis is performed to assess the discriminative power of the learned latent space features. The latent features of one representative male and female subject are shown in Figure~\ref{fig:graph}, clearly highlighting the significant difference in SVAE latent space pattern between male and female.
The SVAE model completed 50 epochs of training, achieving an average composite loss of approximately 1.8126. Training stability is maintained through a reduced learning rate at $10^{-4}$ and gradient clipping, preventing any NaN or Inf loss values.

Using 10-fold stratified cross-validation on the SVAE-derived latent features, a comprehensive benchmark is conducted across $12$ individual ML classifiers. The features are standardized within each fold before training. The mean performance metrics across all folds are summarized in Table~\ref{tab:table1}.

\begin{figure}[btp]
    \centering
    \includegraphics[width=0.9\linewidth]{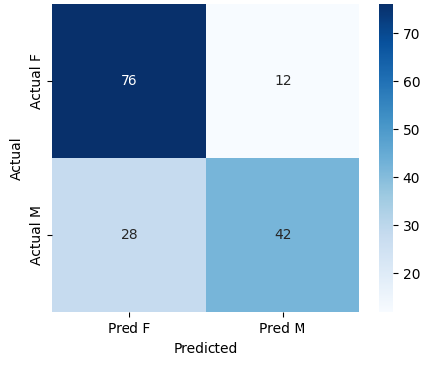}
    \caption{The heat map of confusion matrix using MLP for sex classification}
    \label{fig:conf-mlp}
\end{figure}

\begin{figure}[btp]
    \centering
    \includegraphics[width=1\linewidth]{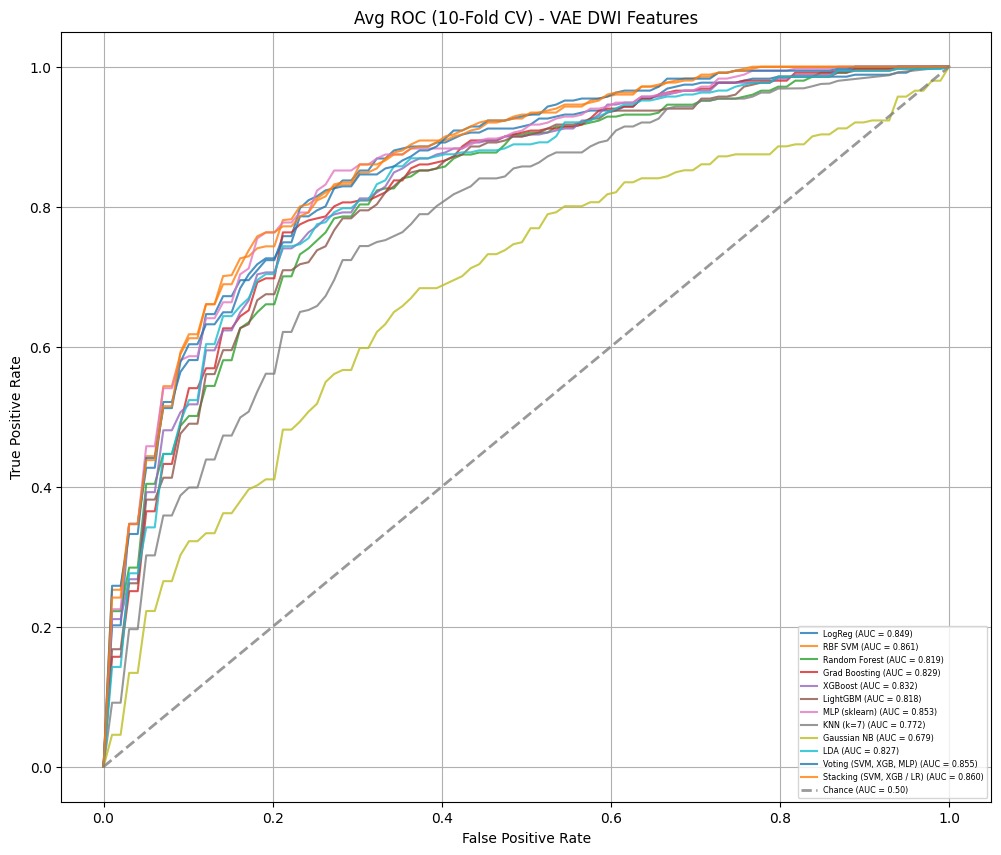}
    \caption{Receiver operating characteristics curves from 10-fold cross-validation using various ML classifiers.}
    \label{fig:roc}
\end{figure}

\begin{figure}[btp]
    \centering
    \includegraphics[width=1\linewidth]{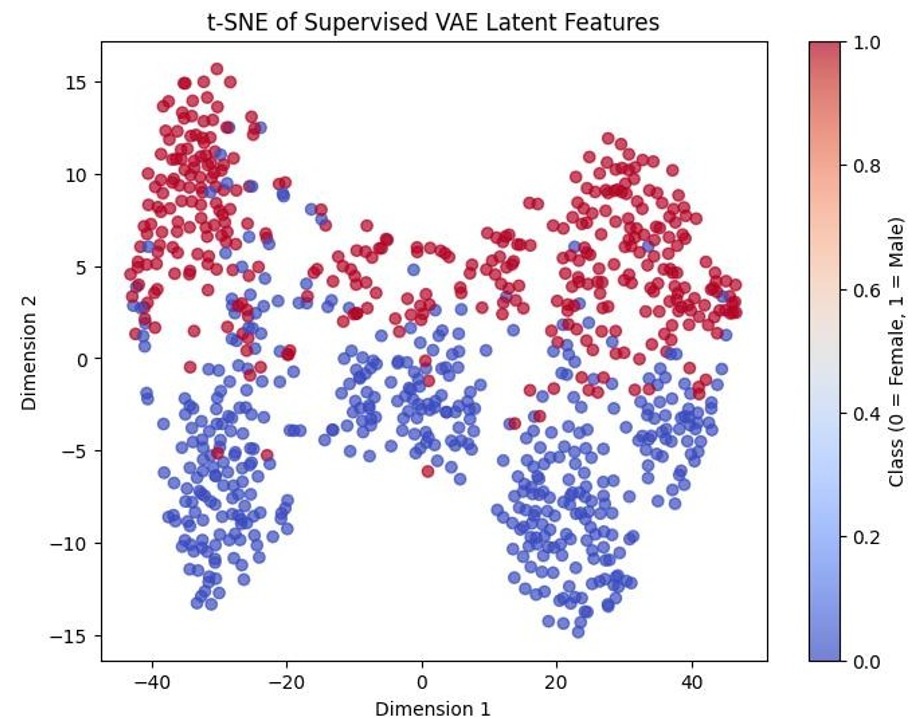}
    \caption{Illustration of sex classification using t-SNE plot; demonstrating the distinct patterns of DWI brain structural connectivity between male and female subjects. This highlights the effectiveness of ML classifiers in distinguishing between the SVAE latent space features of males and females subjects, considered in this study.}
    \label{fig:tsne}
\end{figure}

\begin{table*}[btp]
\centering
\small 
\setlength{\tabcolsep}{6pt} 
\caption{\centering Summary of sex classification performance measures using SVAE latent space features. Among all ML classifier MLP outperformes with 80\% accuracy.}
\label{tab:table1}
\begin{tabular}{|c|c|c|c|c|c|} 
\hline
\textbf{Model} & \textbf{Precision} &\textbf{Recall} &\textbf{F1-score} &\textbf{Accuracy} & \textbf{AUC} \\
\hline
Stacking (SVM,XGB/LR) & 0.749 ($\pm0.038$) & 0.774 ($\pm0.055$) & 0.781 ($\pm0.035$) & 0.784 ($\pm$0.035) & 0.860 ($\pm$0.032) \\
\hline
RBF SVM & 0.747 ($\pm0.028$) & 0.769 ($\pm0.056$) & 0.779 ($\pm0.030$) & 0.781 ($\pm$0.030) & 0.861 ($\pm$0.032) \\
\hline
Logistic Regression & 0.741 ($\pm0.034$) & 0.777 ($\pm0.066$) & 0.778 ($\pm0.038$) & 0.780 ($\pm$0.038) & 0.849 ($\pm$0.029) \\
\hline
Voting (SVM, XGB, MLP) & 0.756 ($\pm0.043$) & 0.712 ($\pm0.058$) & 0.765 ($\pm0.040$) & 0.770 ($\pm$0.039) & 0.855 ($\pm$0.033) \\
\hline
LDA & 0.747 ($\pm0.046$) & 0.723 ($\pm0.055$) & 0.763 ($\pm0.041$)  & 0.768 ($\pm$0.041) & 0.827 ($\pm$0.025) \\
\hline
Gradient Boosting & 0.744 ($\pm0.043$) & 0.709 ($\pm0.068$) & 0.758 ($\pm0.045$) & 0.762 ($\pm$0.043) & 0.829 ($\pm$0.046) \\
\hline
XGBoost & 0.732 ($\pm0.033$)& 0.723 ($\pm0.078$) & 0.756 ($\pm0.041$) & 0.760 ($\pm$0.040) & 0.832 ($\pm$0.042) \\
\hline
LightGBM & 0.727 ($\pm0.059$) & 0.694 ($\pm0.070$) & 0.742 ($\pm0.051$) & 0.747 ($\pm$0.051) & 0.818 ($\pm$0.054) \\
\hline
Random Forest & 0.723 ($\pm0.050$) & 0.660 ($\pm0.070$) & 0.729 ($\pm0.042$) & 0.736 ($\pm$0.041) & 0.819 ($\pm$0.037) \\
\hline
KNN (k=7) & 0.711 ($\pm0.074$) & 0.546 ($\pm0.083$) & 0.683 ($\pm0.056$) & 0.699 ($\pm$0.052) & 0.772 ($\pm$0.058) \\
\hline
Gaussian NB & 0.519 ($\pm0.077$) & 0.720 ($\pm0.159$) & 0.519 ($\pm0.094$) & 0.548 ($\pm$0.079) & 0.679 ($\pm$0.056) \\
\hline
MLP  & 0.787 ($\pm0.044$) & 0.723 ($\pm0.081$) & 0.783 ($\pm0.041$) & \textbf{0.80 ($\pm$0.039)} & \textbf{0.853 ($\pm$0.034)} \\
\hline

\end{tabular}
\end{table*}

Among all classifiers, the MLP classifier demonstrated the highest performance on SVAE-derived features, achieving a mean accuracy of 0.80 $\pm$ 0.0389 (Figure~\ref{fig:conf-mlp}. Followed by MLP, the similar accuracy values are obtained using the stacking classifier (0.78 $\pm$ 0.03 accuracy, 0.86 $\pm$ 0.03 AUC) and RBF SVM (0.78 $\pm$ 0.03 accuracy, 0.86 $\pm$ 0.03 AUC). Logistic Regression also performed comparably in accuracy (0.78 $\pm$ 0.04).
In ensemble methods, the stacking classifier yield best accuracy, highlighting the benefit of model combination. Tree-based models like XGBoost (0.76 accuracy), Gradient Boosting (0.76 accuracy), and LightGBM (0.74 accuracy) offered competitive performance, while simpler models such as KNN and Naive Bayes are less effective, with Naive Bayes showing the lowest performance. The ROC curves, as shown in Figure~\ref{fig:roc} visually corroborated these findings, with MLP, Stacking, RBF SVM, and Logistic Regression exhibiting superior performance, and RBF SVM and the stacking classifier achieving the highest mean AUC values. The t-SNE plot of the result obtained from SVAE is shown in Figure~\ref{fig:tsne}. 
To better understand the contribution of the SVAE, we conducted a baseline experiment using a simple CNN model trained on raw connectome features. However, our approach outperforms this baseline analysis, achieving a 3\% improvement in performance.
This highlights the ability of SVAE to learn compact, informative representations that capture sex-specific characteristics from complex DWI data.

The present study investigates whether the structural connectivity among different brain regions, captured by DWI connectomes, could reliably differentiate sex in young individuals. To achieve this, SVAE framework is employed for deriving latent space features of DWI connectome, which are subsequently fed into various ML models to assess the efficacy of these latent representations for sex classification problem. Our experiments revealed a maximum classification accuracy of 80\% using MLP, underscoring the potential of this approach.
A key strength of our study lies in the use of a large sample of 787 young adults, which enhances the generalizability of our findings. While most of the previous studies investigated differences in structural anatomy or function  \cite{pnas,b22,b23,b24}, this study emphasizes the utility of SVAE for DWI connectome based sex classification.
Our results demonstrate that the proposed SVAE framework effectively captures latent features of structural brain connectivity that are informative for distinguishing between sexes. This finding highlights the strength of our approach in learning discriminative patterns from the DWI connectome - representing a significant step as compared to previous approaches, reported in literature. The present work highlights the potential of using DWI brain connectomics and deep learning to study sex-related brain differences, paving the way for future research on diverse populations that may include transgender individuals, to better understand sex-related variations in brain structure.

While this study successfully demonstrates the utility of a supervised VAE (SVAE) framework for sex classification, several limitations should be acknowledged. Firstly, although the study utilized a substantial cohort of nearly 787 subjects, the generalizability of the findings could be enhanced by validation on larger and more diverse datasets, including individuals from different age ranges, ethnicities, and clinical populations. A more heterogeneous dataset would provide a more rigorous test of the model's robustness and its ability to learn fundamental sex-specific patterns independent of sample-specific characteristics. Secondly, the quality of the SVAE's generative capability was primarily evaluated through the lens of classification performance rather than a detailed analysis of the reconstruction loss or the visual fidelity of the reconstructed connectome matrices. The composite loss function, which balanced reconstruction, KL divergence, and classification objectives, may have prioritized discriminability over perfect reconstruction, and a different weighting scheme could yield different latent space characteristics. Finally, this work utilized a fully connected architecture for the SVAE; exploring convolutional VAEs could potentially offer better preservation of the inherent spatial topology of the connectivity matrices, though initial experiments with other deep learning architectures showed challenges. These aspects represent important avenues for future research.

\section{Conclusion}
\label{conclusion}

The present study uses SVAE framework from DWI structural brain connectivity matrix for the task of sex classification. The present study can be extended to apply similar techniques to analyze the connectomes of transgender individuals, which could provide valuable insights into the neural correlates of gender identity and diversity. Additionally, future work could incorporate longitudinal analyses to explore how connectome patterns evolve across different developmental stages or under the influence of hormonal or medical interventions. Furthermore, integrating multimodal medical data will aid to develop a more holistic understanding of sex differences and gender diversity in brain connectivity. Finally, exploring model interpretability, and generalizability across diverse populations will be essential for conclusive inferences.


\bibliographystyle{IEEEtran}
\bibliography{SVAE_conference}


\end{document}